# On Integrating Fuzzy Knowledge Using a Novel Evolutionary Algorithm

Nafisa Afrin Chowdhury, Murshida Khatun and M. M. A. Hashem

Department of computer science and Engineering,

Khulna University of Engineering & Technology, Khulna

Khulna-920300, Bangladesh

Ph: +88041-774318, Fax+88-0141774403

E-mail: nafisaafrin@gmail.com;

mma_hashem@yahoo.com.

**Abstract**

*Fuzzy systems may be considered as knowledge-based systems that incorporates human knowledge into their knowledge base through fuzzy rules and fuzzy membership functions. The intent of this study is to present a fuzzy knowledge integration framework using a Novel Evolutionary Strategy (NES), which can simultaneously integrate multiple fuzzy rule sets and their membership function sets. The proposed approach consists of two phases: fuzzy knowledge encoding and fuzzy knowledge integration Four application domains, the hepatitis diagnosis, the sugarcane breeding prediction, Iris plants classification, and Tic-tac-toe endgame were used to show the performance of the proposed knowledge approach. Results show that the fuzzy knowledge base derived using our approach performs better than Genetic Algorithm based approach.*

## 1. Introduction

The construction of a reliable knowledge-based system, especially for complex application problems, usually requires the integration of multiple knowledge inputs as related domain knowledge is usually distributed among multiple sites. The use of knowledge integrated from multiple knowledge sources is thus important to ensure comprehensive coverage. Most knowledge sources or actual instances in real-world applications, especially in medical and control domains, contain fuzzy or ambiguous information. Expressions of domain knowledge in fuzzy terms are thus seen more and more frequently. Several researchers have recently investigated automatic generation of fuzzy classification rules and fuzzy membership functions using evolutionary algorithms [1].

Many knowledge acquisition and integration systems [4] use genetic algorithms to derive knowledge from training instances. In this study we have proposed an approach that uses a novel evolutionary algorithm for fuzzy knowledge integration process. In this technique, the emphasis is not on the acquisition of a structure with high fitness, as in GA, but on modeling evolution at the granularity of an individual, ESs considers an individual to be composed of a set of behaviors, each of which is a feature. In this approach we used insertion deletion mutation as a variation (genetic) operator in order to introduce variable length string for each individual. Therefore the integration process shows fast convergence with considerable accuracy as comparing with GA approach.

The remainder of this paper is organized as follows. A NES based fuzzy knowledge integration framework is proposed in **section 2**. The fuzzy knowledge encoding approach is explained in **section 3.** The fuzzy knowledge integration approach is proposed in **section 4.** Experiments on two application domains- Hepatitis Diagnosis and Sugarcane Breeding Prediction are stated in **section 5**. Conclusion and future research directions are given in **section 6.**

2. **Fuzzy Knowledge Integration Framework**

Here, we propose a NES-based fuzzy-knowledge integration framework. The proposed framework can integrate multiple fuzzy rule sets and membership function sets at the same time [4]. Fuzzy rule sets, membership functions, and test objects including instances and historical records may be distributed among various sources. Knowledge from each site might be directly obtained by a group of human experts using knowledge-acquisition tools or derived using machine-learning methods. Here, we assume that all knowledge sources are represented by fuzzy rules since almost all knowledge derived by knowledge-acquisition tools or induced by machine-learning methods may easily be translated into or represented by rules.

The proposed framework maintains a population of fuzzy rule sets with their membership functions and uses a Novel Evolutionary Algorithm to automatically derive the resulting fuzzy knowledge base. This integration framework operates in two phases: fuzzy knowledge encoding and fuzzy knowledge integration. The encoding phase first transforms each fuzzy rule set and its associated membership functions into an intermediary representation, which is further encoded as a variable-length string [4]. The integration phase then chooses appropriate strings for "mating," gradually creating good offspring fuzzy rule sets and membership function sets. The offspring fuzzy rule sets with their associated membership functions then undergo recursive "evolution" until an optimal or nearly optimal set of fuzzy rules and membership functions has been obtained. Fig. 1 shows the two-phase process, where are the fuzzy rule sets with their associated membership function sets, as obtained from different sources for integration.

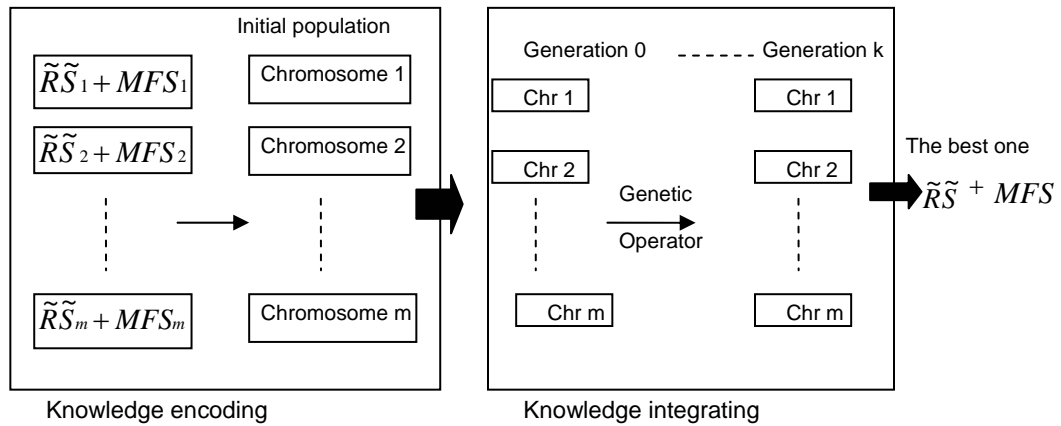

Fig. 1: The genetic-fuzzy knowledge integration process.

3. **Knowledge Encoding**

The encoding phase first transforms each fuzzy rule set and its associated membership functions into an intermediary representation in a few steps [4], further encoded as a variable-length string. Fig 2 shows μ individuals constructing a generation. Each individual has rule set and membership functions that represent the feature values and corresponding membership criteria for each feature respectively.

|Ind 1|Rule 1|Rule 2|----|MF 1|MF 2|----|
|Ind 2|Rule 1|Rule 2|----|MF 1|MF 2|----|
|⋮|||||||
|Ind μ|Rule 1|Rule 2|----|MF 1|MF 2|----|

Fig.2: The rule set and membership function encoding of a generation having μ individuals.

To effectively encode the associated membership functions, we use two parameters to represent each membership function, as Parodi and Bonelli [4] did. Membership functions applied to a fuzzy rule set are then assumed to be isosceles-triangle functions as shown in Fig. 3, where $a_{ij}$ denotes the jth linguistic value of Feature $A_i$, $u_{a_{ij}}$ denotes the membership function of $a_{ij}$, $c_{ij}$ indicates the center abscissa of membership function $u_{a_{ij}}$ and $w_{ij}$ represents half the spread of membership function $u_{a_{ij}}$.

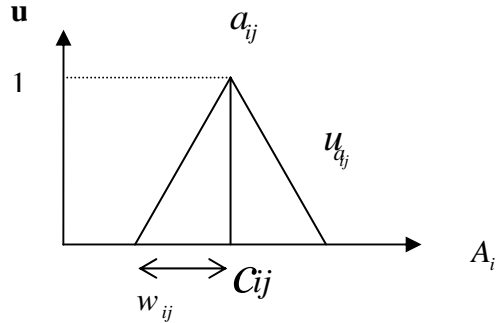

Fig.3: Membership functions of feature $A_i$.

### 4. Knowledge Integration

After each fuzzy rule set with its associated membership functions has been encoded as a string, the fuzzy-knowledge integration process starts. It chooses good individuals in the population for "mating," gradually creating better offspring fuzzy rule sets and membership function sets. In order to make the integration process more effective and to increase accuracy of the resultant knowledge base we used a Novel Evolutionary Strategy (NES) Algorithm in place of Genetic Algorithm (GA). There are μ individuals are created initially by the program using the function urnN_N (a, b) for input variables and output variables. Their corresponding evaluated fitness $f$ value for each individual is also calculated in the evaluated function. Two important factors are used in evaluating process. They are: accuracy (*RS*) and complexity (*RS*) of the resulting knowledge structure. Accuracy of a fuzzy rule set $\widetilde{R}\widetilde{S}$ is evaluated using test objects as follows [4]:

$$\text{Accuracy}(\widetilde{R}\widetilde{S}) = \frac{\text{total number of objects correctly matched by } \widetilde{R}\widetilde{S}}{\text{total number of objects}} \quad \ldots\ldots\ldots\ldots\ldots 1$$

The more data used the more objective and accurate the evaluation is. The complexity of the resulting rule set $(\widetilde{R}\widetilde{S})$ is the ratio of rule increase, defined as follows:

$$\text{Complexity}(\widetilde{R}\widetilde{S}) = \frac{\text{Number of rules in the integrated rule set } \widetilde{R}\widetilde{S}}{[\sum_{i=1}^{p}(\text{Number of rules in the initial } \widetilde{R}\widetilde{S}_i)]/p} \quad \ldots\ldots\ldots\ldots 2$$

Where $\widetilde{R}\widetilde{S}_i$ is the *i*th initial fuzzy rule set, and *P* is the number of initial rule sets. Accuracy and complexity are combined to represent the fitness value of the rule set. Since the goal is to increase the accuracy and decrease the complexity of the resulting rule sets, the evaluation function is then defined as follows:

$$\text{Fitness}(RS) = \frac{[\text{Accuracy}(RS)]}{[\text{Complexity}(RS)]^{\alpha}} \quad \ldots\ldots\ldots\ldots\ldots\ldots\ldots\ldots\ldots\ldots\ldots\ldots 3$$

Where $\alpha$ is a control parameter, representing a tradeoff between accuracy and complexity.

Two genetic operators, crossover and mutation, are applied on individuals of each generation to create new offspring.

## A. SBMAC

In the SBMAC (Sub-population Based Max-mean Arithmetical Crossover) each subpopulation's elite and the mean –individual (the virtual parent) created from that subpopulation excluding the elite are used. This SBMAC is supposed to explore promising areas in the search space with different directions towards the optimum point [3]. Thus, the algorithm is exposed to less possibility of being trapped in local optima.

## B. Modified SBMAC

For an instance of the evolution, a modified version of the SBMAC operation is shown in fig 4. Firstly, the minimum length individual within each subpopulation is determined. Secondly, the offspring variables are produced up to this length [5].Then the remaining part of the parent subpopulation variables is appended to the offspring population variables to form the full length chromosome.

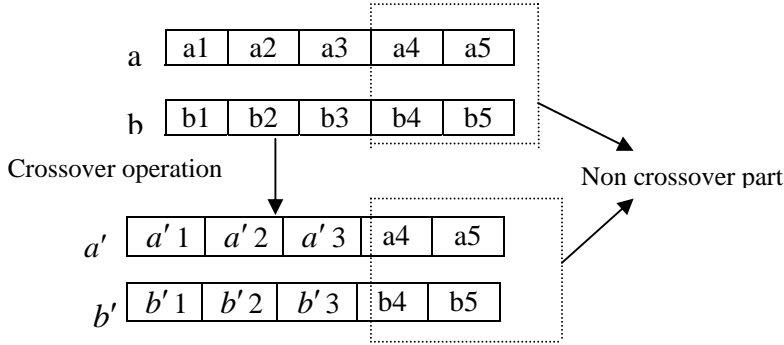

Fig 4: Modified SBMAC operation.

## C. TVM

In the real world, a rapid change is observed at the early stages of life and a slow change is observed at latter stages of life in all kind of animals/plants. These changes are more often occurred dynamically depending on the situation exposed to them. A special dynamic Time-Variant Mutation (TVM), which follows this natural evidence [3], is applied here.

## D. Insertion Deletion Mutation

In order to generate variable length chromosome for each individual insertion deletion mutation [5] is performed on each individual. For insertion mutation, a new randomly generated steering angle value (gene) is inserted at a selected position based on a probability of insertion $P_{im}$. This will increase the length of a chromosome. On the other hand the deletion mutation is the reverse action than that of the insertion mutation. Fig 5 shows the insertion deletion mutation procedure.

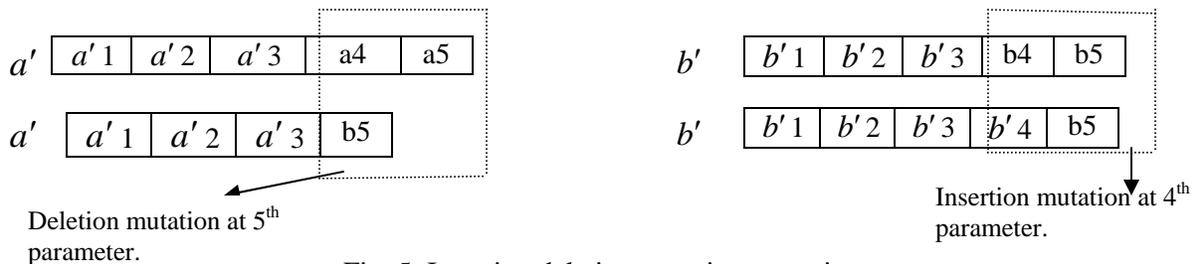

Fig. 5: Insertion deletion mutation operation.

## 5. Simulation result

To demonstrate the effectiveness of the proposed fuzzy knowledge integration approach, we applied it to four application domains. The first one was "The Hepatitis Diagnosis Domain", which contained 155 cases obtained from Carnegie-Mellon University [4]. The second was "The Sugarcane Breeding Prediction Domain", which contained 699 cases obtained from Taiwan Sugar Research Institute (TSRI) [4]. The flexibility of this approach is proved by applying successfully in two new application domains, where the GA approach is not applied yet. They are: "The Iris Plants Classification Domain" [7] and "The Tic- Tac-Toe Endgame Domain" [7].

### 5.1 The Hepatitis Diagnosis Domain:

The hepatitis diagnosis problem was first used here to test the performance of the proposed fuzzy-knowledge integration approach. The goal of the experiments was to identify two possible classes, *Die* or *Live*, from a set of instances. Table I shows an actual case expressed in term of 19 features and one class. Among these features, Age, Bilirubin, Alk Phosphate, SGOT, Albumin, and Protime are numerical and have membership functions with them.

**Table I**
**A case for hepatitis diagnosis**

| Features | Features value |
|---|---|
| Billirubin | 0.90 |
| Alk Phosphate | 95 |
| SGOT | 28 |
| Albumin | 4.0 |
| Protime | 75 |

**Class: Live**

Each rule, consisting of 5 feature tests and a class pattern, was encoded into a record having 5 rule values and 10 membership function value (center, width). The parameter $\alpha$ was set at 0.01. Experimental results show that executing the proposed approach over more generations initially yielded more accurate results and finally gradually converged to a constant. The structure of the best individual after convergence is shown in fig.6. Fig. 6(a), 6(b), 6(c), 6(d), 6(e) are for five numerical parameters- Billirubin, Alk Phosphate, SGOT, Albumin, Protime.

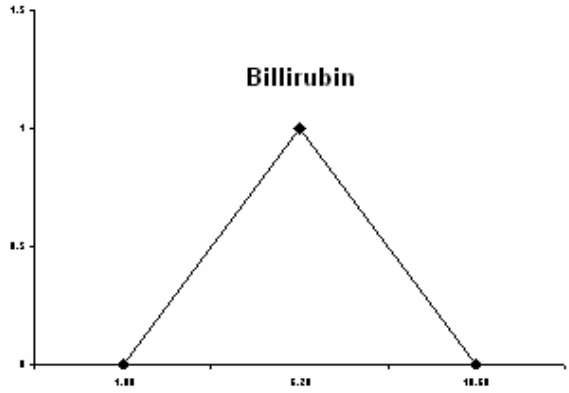

Fig. 6 (a): Membership function of Billirubin.

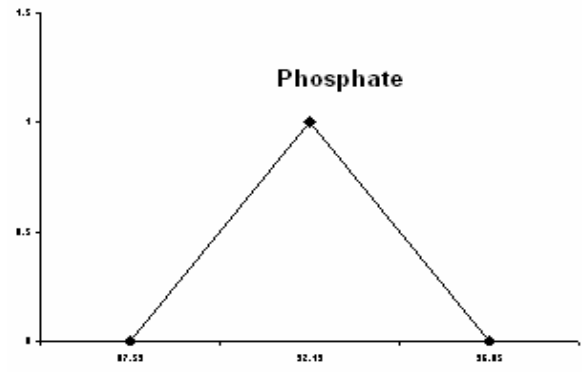

Fig. 6 (b): Membership function of Alk phosphate.

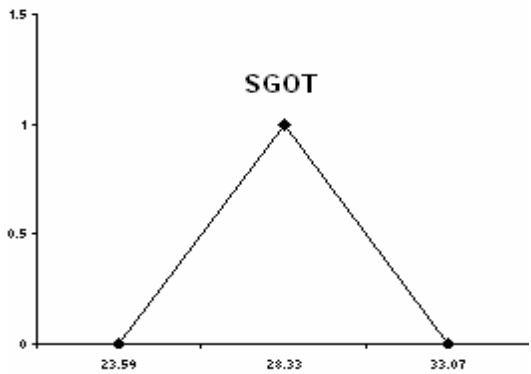

Fig. 6 (c): Membership function of SGOT.

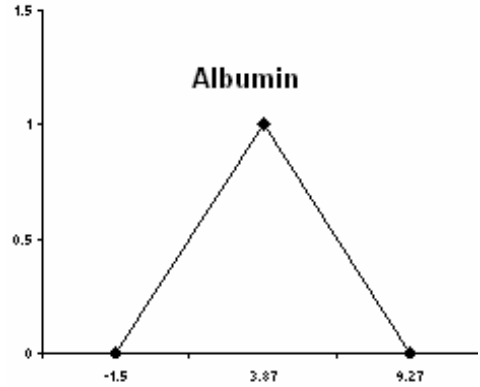

Fig. 6 (d): Membership function of Albumin.

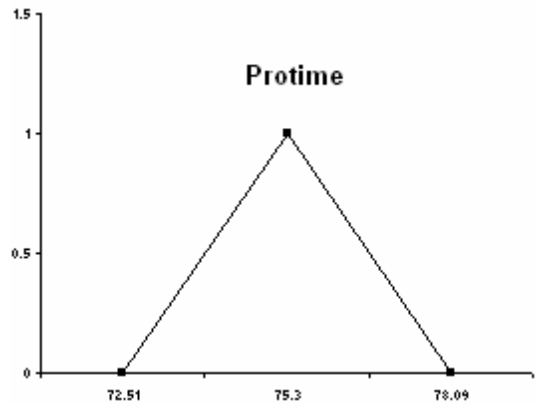

Fig. 6 (e): Membership function of Protime.

Fig 7 shows the relationship between the accuracy of the resulting rule sets with the number of generations by the proposed approach. As the numbers of generations increased, the accuracy as well as resulting fitness value also increased and finally converged to a specific value.

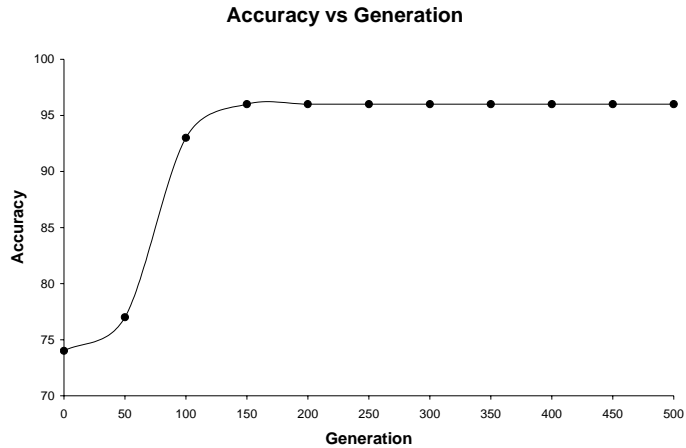

Fig.7: Relationship between accuracy and number of generations for the hepatitis domain

Table II compares the accuracy of our proposed approach with that of the above learning methods [4]. It can easily be seen that our approach has a higher accuracy than some other learning methods. Moreover, in our approach convergence occurs after 200 generation whereas in GA approach, it takes 4000 generation [4].

**Table II**
A comparison with some other learning methods for hepatitis domain

| Methods | Accuracy |
|---|---|
| Our approach | 96.33% |
| GA approach | 91.61% |

*5.2 The Sugarcane Breeding Prediction Domain*

In this experiment, we present a real application to the sugarcane breeding prediction [4] based on our proposed approach. The goal of this application was to breed good sugarcane plants with high sugar contents and strong disease-resistance abilities.

**Table III**
A Case for Sugarcane Breeding Prediction

| Features | Values |
|---|---|
| Stalk diameter | 3.5cm |
| Stalk length | 5.2m |
| Stalk number | 96 |
| Cane yield | 198 |

**Offspring with high sugar-ingredients and disease-resistant abilities**

Table III shows an actual case expressed in terms of 36 features and a class report. Here, the parameter $\alpha$ was set at 0.001. Our approach obtained an accuracy rate of 89.52% after 500 execution generations. In fig. 8 the resultant structure of the best individual is represented. Fig 8 (a), 8 (b), 8 (c), 8 (d) shows the resultant membership function for Stalk diameter, Stalk length, Stalk number and Cane yield, respectively.

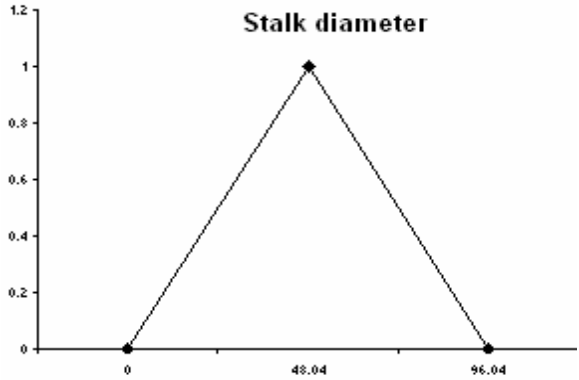

Fig 8 (a): Membership function of Stalk diameter.

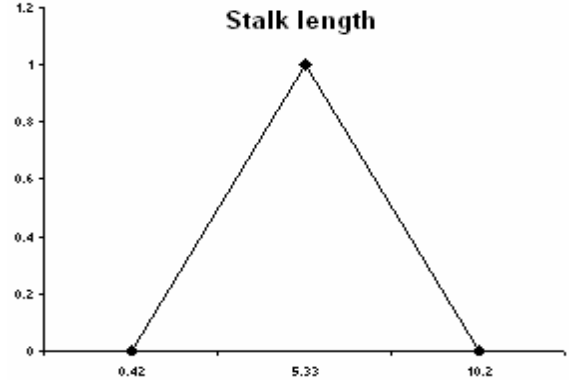

Fig 8 (b): Membership function of Stalk length.

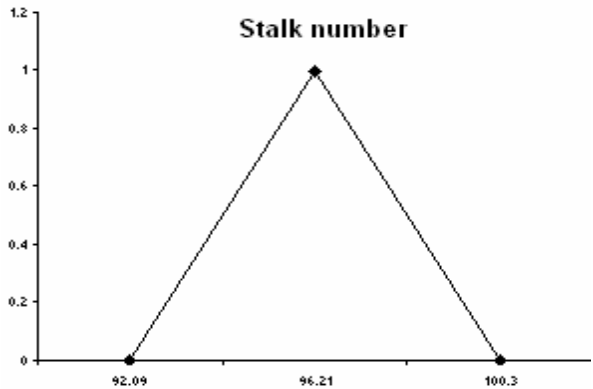

Fig 8 (c): Membership function of Stalk number.

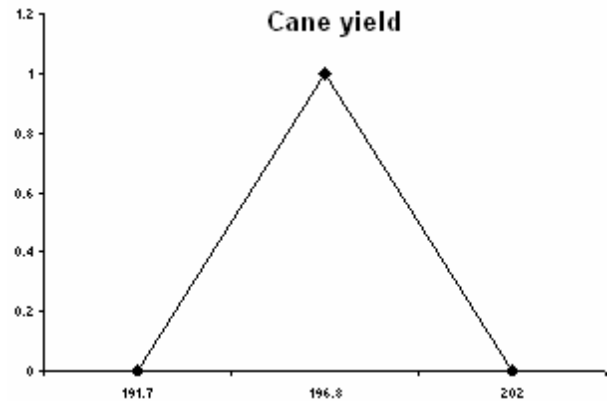

Fig 8 (d): Membership function of Cane yield.

Fig.9 shows the results for different generations with respect to accuracy in the sugarcane domain. Table IV compares the accuracy of our proposed approach with that of the sugarcane breeding assistant system (SCBAS) learning method [4] in the sugarcane domain. It can easily be seen that our approach has a higher accuracy than the GA approach or SCBAS learning method. Moreover, in our approach convergence occurs after 400 generation whereas in GA approach, it takes 4500 generation[4].

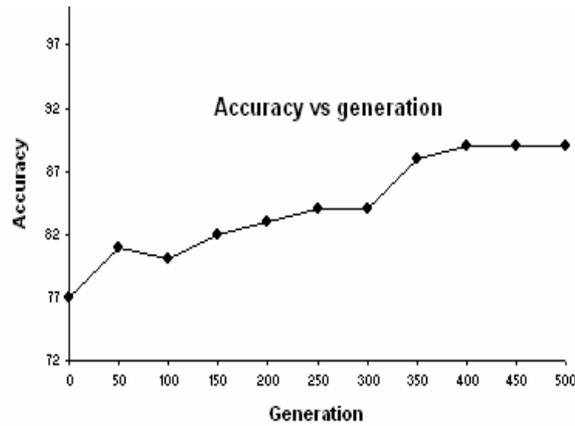
Fig 9: Relationship between accuracy and generation for the sugarcane domain.
**Table IV**
A comparison with some other learning methods for sugarcane breeding prediction domain.

| Methods | Accuracy |
|---|---|
| Our approach | 89.01% |
| GA approach | 76.02% |

In case of Iris plants classification our approach shows the accuracy of 76% after 100 generation and it is 67% after 300 generation in case of Tic-tac-toe endgame.

## 6. Concluding Remarks

In this paper, we have proposed an appropriate representation to encode the fuzzy knowledge and have shown how fuzzy-knowledge integration can be effectively processed using this representation. Experimental results have also shown that our genetic fuzzy-knowledge integration framework is valuable for simultaneously combining multiple fuzzy rule sets and membership function sets. Our approach needs no human experts' intervention during the integration process. The time required by our approach is thus dependent on computer execution speed, but not on human experts. Much time can thus be saved since experts may be geographically dispersed. Also, our approach is a scalable integration method that can be applied as well when the number of rule sets to be integrated increases. Integrating a large number of rule sets may increase the validity of the resulting knowledge base. It is also objective since human experts are not involved in the integration process.